\documentclass[10pt,twocolumn,letterpaper]{article}

\usepackage{cvpr}
\usepackage{times}
\usepackage{epsfig}
\usepackage{graphicx}
\usepackage{amsmath}
\usepackage{amssymb}

\usepackage[T1]{fontenc}
\usepackage[utf8]{inputenc}
\usepackage{authblk}

\usepackage[breaklinks=true,bookmarks=false]{hyperref}

\cvprfinalcopy 

\def\cvprPaperID{****} 

\begin{document}


\makeatletter
\renewcommand\AB@affilsepx{, \protect\Affilfont}
\makeatother

\title{\vspace{-.12cm}Visual to Sound: Generating Natural Sound for Videos in the Wild}
\author[1]{\vspace{-.6cm}Yipin Zhou}
\author[2]{Zhaowen Wang}
\author[2]{Chen Fang}
\author[2]{Trung Bui}
\author[1]{Tamara L. Berg}
\affil[1]{University of North Carolina at Chapel Hill}
\affil[2]{Adobe Research\vspace{-.4cm}}


\maketitle
\begin{abstract}
As two of the five traditional human senses (sight, hearing, taste, smell, and touch), vision and sound are basic sources through which humans understand the world. Often correlated during natural events, these two modalities combine to jointly affect human perception. In this paper, we pose the task of generating sound given visual input. Such capabilities could help enable applications in virtual reality (generating sound for virtual scenes automatically) or provide additional accessibility to images or videos for people with visual impairments. As a first step in this direction, we apply learning-based methods to generate raw waveform samples given input video frames. We evaluate our models on a dataset of videos containing a variety of sounds (such as ambient sounds and sounds from people/animals). Our experiments show that the generated sounds are fairly realistic and have good temporal synchronization with the visual inputs.
\end{abstract}
\vspace{-.4cm}

\section{Introduction}
\label{sec:intro}
The visual and auditory senses are arguably the most important channels through which humans perceive their surrounding environments, and they are often entertwined. From life-long observations of the natural world, people are able to learn the association between vision and sound. For instance, when seeing a flash of lightning in the sky, one might cover their ears subconsciously, knowing that the crack of thunder is coming next. Alternatively, hearing leaves rustling in the wind might conjure up a picture of a peaceful forest scene.

In this paper, we explore whether computational models can learn the relationship between visuals and sound. Models of this relationship could be fundamental for many applications such as combining videos with automatically generated ambient sound to enhance the experience of immersion in virtual reality; adding sound effects to videos automatically to reduce tedious manual sound editing work; Or enabling equal accessibility by associating sound with visual information for people with visual impairments (allowing them to ``see'' the world through sound). While all of these tasks require powerful high-level inference and reasoning ability, in this work we take a first step toward this goal, narrowing down the task to generating audio for video based on the viewable content. 

Specifically, we train models to directly predict raw audio signals (waveform samples) from input videos. The models are expected to learn associations between generated sound and visual inputs for various scenes and object interactions. Existing works~\cite{visual_indicated_sound,v2s2v} handle sound generation given input of videos/images under experimental settings (e.g., to generate a hitting sound or where the input videos are recorded indoor with fixed background). In our work, we deal with generating natural sound from videos collected in the wild.

To enable learning, we introduce a dataset that is derived from AudioSet~\cite{audioset}. AudioSet is a dataset collected for audio event recognition but not ideal for our task because many of videos and audios are loosely related; the target sound might be covered by other sounds (like music); and the dataset contains some mis-classified videos. All of these sources of noise tend to deter the models from learning the correct mapping from video to audio. To alleviate these issues, we clean a subset of the data, including sounds of humans/animals and other natural sounds, by verifying the presence of the target objects for both videos and audios respectively (at 2 second intervals) to make them suitable for the generation task (Sec.~\ref{sec:dataset}). 

Our model learns a mapping from video frames to audio using a video encoder plus sound generator structure. For sound generation, we use a hierarchical recurrent neural network proposed by~\cite{SampleRNN}. We present 3 variants to encode the visual information, which can be combined with the sound generation network to form a complete framework (Sec.~\ref{sec:methods}). To evaluate the proposed models and the generated results, we conduct both numerical evaluations and human experiments (Sec.~\ref{sec:experiments}). Please see our supplementary video to see and hear sound generation results. 

The innovations introduced by our paper are: 1) We propose a new problem of generating sounds from videos in the wild; 2) We release a dataset containing 28109 cleaned videos (55 hours in total) spanning 10 object categories; 3) We explore model variants for the generation architectures; 4) Numerical and human evaluations are provided as well as an analysis of generated results.

\section{Related work}
\label{sec:relatedwork}
\noindent {\bf Video and sound self-supervision:} The observation that audio and video may provide supervision for each other has drawn attention recently. \cite{look,ambient,soundnet,spoken} make use of the concurrent property of video and sound as the supervision to train a network using unlabeled data. \cite{look} presents a two-stream neural network which takes video frames and an audio as inputs to determine whether there is a correspondence (from one video) or not. The network is able to learn both visual and sound semantics through unlabeled videos in a unsupervised manner. Similarly, \cite{spoken} predicts similarity scores for input images and spoken audio spectrum to understand captions based on visual supervision. \cite{soundnet} trains a network to embed visual and audio to learn a deep representation of natural sound without ground truth labels. And \cite{ambient} predicts sound based on associated video frames, instead of generating sound, the goal is to learn visual feature with the guidance of sound clustering.   
%
%

\vspace{+.1cm}
\noindent {\bf Speech synthesis:} The task of speech synthesis is to generate human speech based on input text. Text to speech (TTS) has been studied for a long time from traditional approaches \cite{hmm_tts,unit_tts,stat_tts,Robust_tts,dnn_tts} to deep learning based approaches \cite{wavenet,SampleRNN,Char2Wav}, among which, WaveNet~\cite{wavenet} has attracted much attention due to the improved generation quality. WaveNet presents a convolutional neural networks with dilation structure to predict new audio digits based on previously generated digits. SampleRNN~\cite{SampleRNN} also achieves appealing results in TTS, proposing a hierarchical recurrent neural network (RNN) to recursively generate raw waveform samples temporally. Its hierarchical RNN structure shows the potential to handle long sequence generation. A follow-up work \cite{Char2Wav} demonstrates a novel Reader-Vocoder model and uses SampleRNN~\cite{SampleRNN} as the vocoder to generate raw speech signals.

\vspace{+.1cm}
\noindent {\bf Mapping visual to sound:} Instead of learning a representation by taking advantage of the natural synchronization property between visual and sound, the goal of \cite{visual_indicated_sound,v2s2v} is to directly generate audio conditioned on input video frames. Specifically, \cite{visual_indicated_sound} predicts hitting sound based on different materials of objects and physical interactions. A dataset, Greatest Hits (human hit/scratch diverse objects using a drum stick), has been collected for this purpose. \cite{v2s2v} proposes two generation tasks Sound-to-Image and Image-to-Sound networks using generative adversarial network \cite{GAN}. The data used to train the models shows subjects playing various musical instruments indoor with a fixed background. Recent work \cite{sound20k} presents a synthesized audio-visual dataset built by physics/graphics/audio engines. Fine-grained attributes  have been controlled for synthesis.
Our work has a similar goal, but differs in that instead of mapping visual to sound under constrained settings, we train neural networks to directly synthesize waveform and handle more diverse and challenging real-world scenarios.


\section{Visually Engaged and Grounded AudioSet (VEGAS)}
\label{sec:dataset}
The goal of this work is to generate realistic sound based on video content and simple object activities. As mentioned in Sec.~\ref{sec:intro}, we do not explicitly handle high-level visual reasoning during sound prediction. For the training videos, we expect visual and sound are directly related (predicting dog sound when seeing a dog) most of the time.

Most existing video datasets~\cite{Youtube8M,sports-1M} include both video and audio channels. However, they are typically intended for visual understanding tasks, thus organized based on visual entities/events. A better choice for us is AudioSet~\cite{audioset}, a large-scale object-centric video dataset organized based on audio events. Its ontology includes events such as fowl, baby crying, engine sounds. Audioset consists of 10-second video clips (with audios) from Youtube. The presence of sounds has been manually verified. But as a dataset designed for audio event detection, AudioSet still cannot perfectly fit our needs because of the following three reasons. First, visual and sound are not necessarily directly related. For instance, sometimes the source of a sound may be out of frame. Second, the target sound might have been largely covered by other noise like background music. Third, mis-classification exists. 

We ran several baseline models using the original data and found that the generated sounds are not clean and often accompanied with other noise like chaotic human chatting. To make the data useful for our generation task, we select a subset of videos from AudioSet and further clean them.
\subsection{Data collection}

\begin{figure*}[t]
\begin{center}
\includegraphics[width=0.8\linewidth]{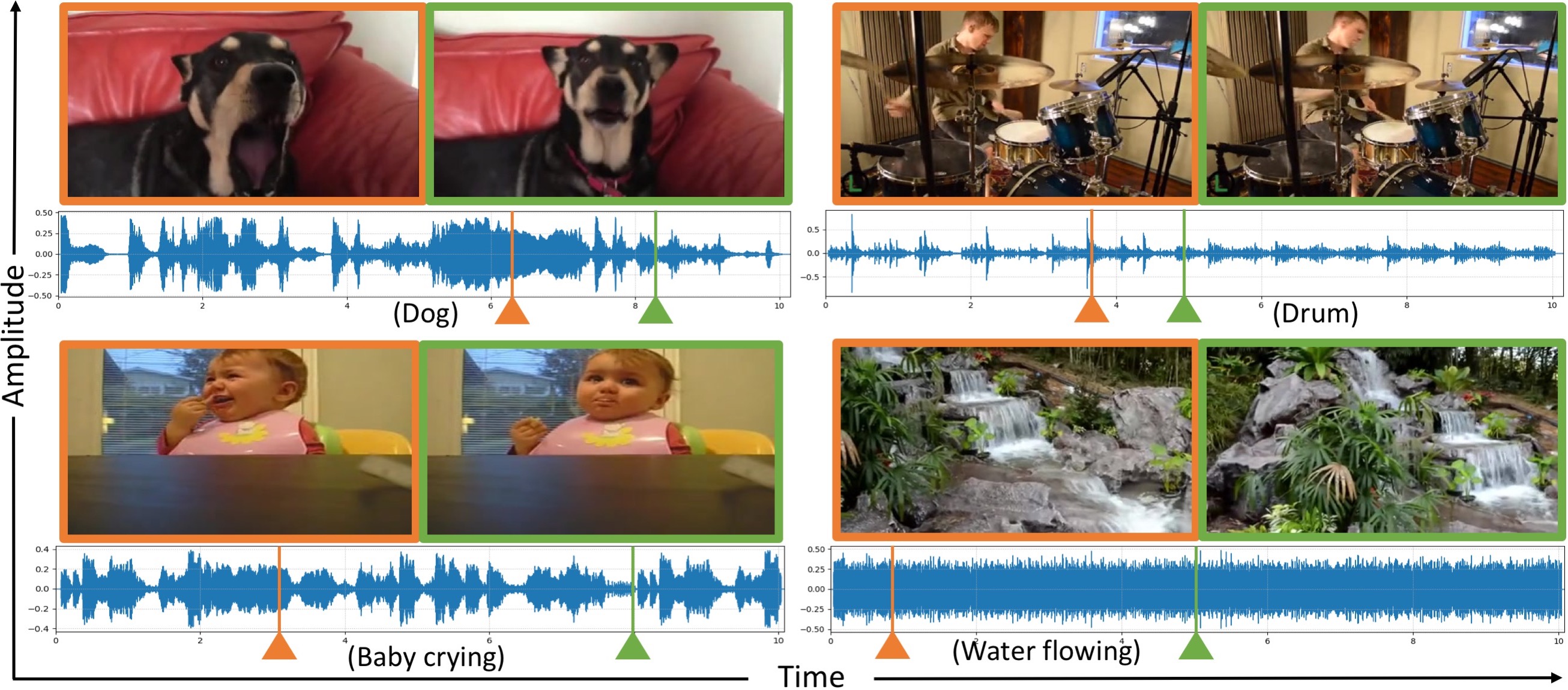}
\end{center}
   \caption{Video frames of 4 categories from the VEGAS dataset with their corresponding waveforms. The color of the image borders is consistent with the mark on the waveform, indicating the position of the current frame in the whole video.}
\vspace{-.2cm}
\label{fig:data_example}
\end{figure*}

We select 10 categories from AudioSet (each including more than 1500 videos) for further cleaning. The selected data includes human/animal sound as well as ambient sounds (specifically they are: Baby crying, Human snoring, Dog, Water flowing, Fireworks, Rail transport, Printers, Drum, Helicopter and Chainsaw). For the categories containing more data than needed, we randomly select 3000 videos for each. 

We use Amazon Mechanical Turk (AMT) for data cleaning, asking turkers to verify the presences of an object/event of interest for the video clip in both the visual and audio modalities. If both modalities are verified we consider it a clean video. 
For most of the videos, noise does not dominate for the entire videos. Therefore, to retain as much data as possible, we segment each video into 2-second short clips for separate labeling. For each short clip, we divide the video and audio for independent annotation.

To clean the audio modality, we ask turkers to annotate the presence of a sound from a target object (e.g. flowing water for water flowing category). Turkers are provided with three choices: 'Yes --- the target sound is dominant over other sounds', 'Sort of --- competitive, 'No --- other sounds are dominant'. To clean the visual modality, we similarly ask turkers to annotate the presence of a target object with three choices: 'Yes --- the target objects appear all the time', 'Sort of --- appearing partially', 'No --- the target objects do not appear or barely appear'. For each segment, we collect annotations from 3 turkers and pick the majority as the final annotation. 
We reject turkers with low accuracy to ensure annotation quality.

We remove the clips where either video and audio has been labeled as 'No' and keep the 'Yes' and Sort of' labeled clips to introduce more variation in the collected data. Finally, we combine the verified adjacent short clips to form longer videos, resulting in videos ranging from 2-10 seconds.

\subsection{Data statistics}
\begin{figure}[t]
\begin{center}
\includegraphics[width=0.8\linewidth]{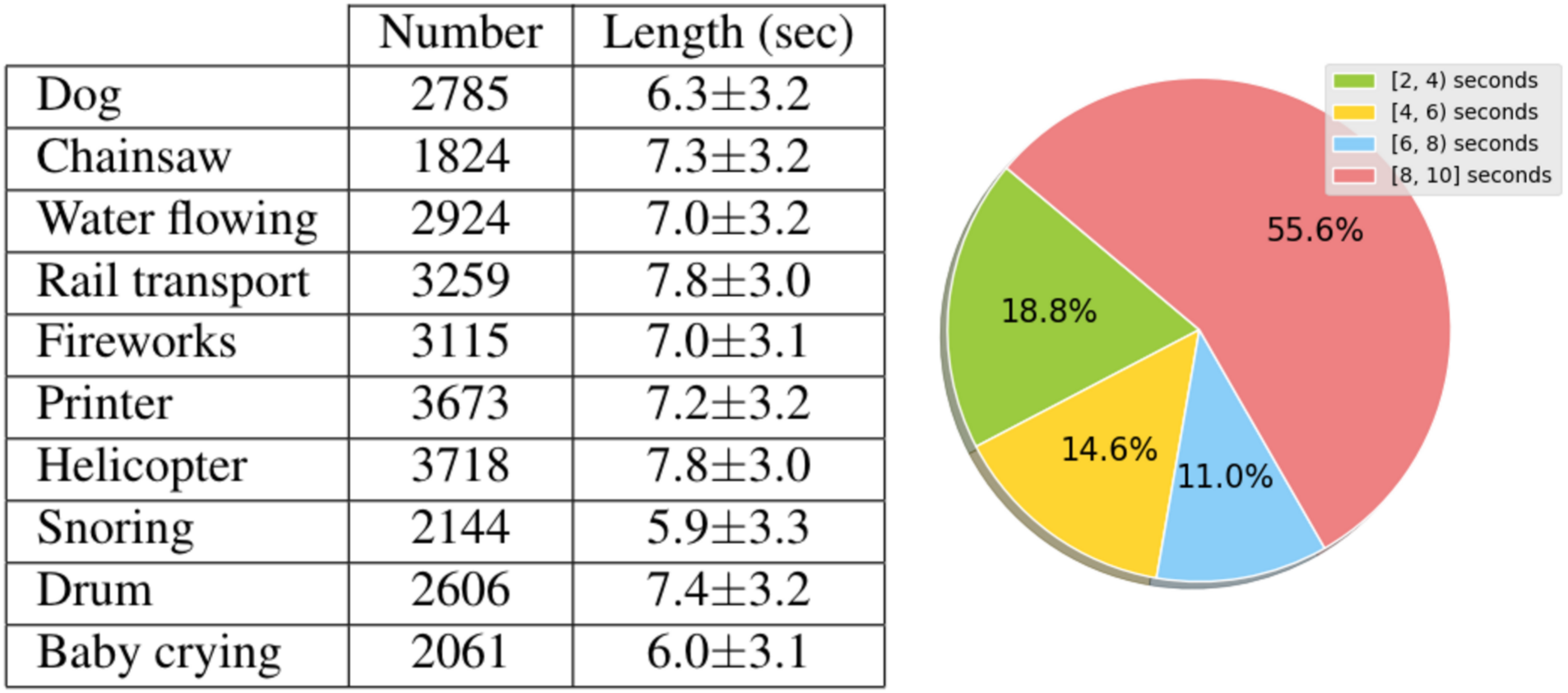}
\end{center}
   \caption{Dataset statistics: the table shows the number of videos with averaged length for each category, while the pie chart presents the distribution of video lengths.}
\vspace{-.3cm}
\label{fig:data_info}
\end{figure}

In total, we annotated 132,209 clips in both the visual and audio modalities, each labeled by 3 turkers, and removed 34,392 clips from the original data. 
After merging adjacent short clips, we have 28,109 videos in total with an average length of 7 seconds and a total length of 55 hours. The left table in Fig.~\ref{fig:data_info} shows the number of videos and the average length with the standard deviation for each category respectively. The pie chart demonstrates the distribution of lengths, showing that the majority of videos are longer than 8 seconds. Fig.~\ref{fig:data_example} shows some example frames with their corresponding waveforms. We can see how sound correlates with the motion of target objects as well as scene events, such as water flowing (bottom right) where the ambient sounds are temporally uniform. Due to the verified properties of the current dataset, we call it the Visually Engaged and Grounded AudioSet (VEGAS).

\section{Approaches}
\label{sec:methods}
In this work, we formulate the task as a conditional generation problem, for which we train a conditional generative model to synthesize raw waveform samples from an input video. Specifically, we estimate the following conditional probability:
\begin{equation}
p(y_1, y_2, ..., y_n | x_1, x_2, ..., x_m)
\end{equation}
where $x_1 ,..., x_m$ represent input video frame representations and $y_1, ..., y_n$ are output waveform values which is a sequence of integers from 0 to 255 (the raw waveform samples are real values ranging from -1 to 1, we rescale and linearly quantize them into 256 bins in our model see Sec.~\ref{sec:sound_generator}). Note that typically $m << n$ because the sampling rate of audio is much higher than that of video, thus the audio waveform sequence is much longer than video frame sequence for a synchronized video.

We adopt an encoder-decoder architecture in model design and experiment with three variants of this type. In general, our models consist of two parts: video encoder and sound generator. In the following sections, we first discuss the sound generator in Sec.~\ref{sec:sound_generator}, then we talk about three different variations of encoding visual information and the concrete systems in Sec.~\ref{sec:frame}, Sec.~\ref{sec:seq} and Sec.~\ref{sec:flow}.

\subsection{Sound generator}
\label{sec:sound_generator}
\begin{figure*}[t]
\begin{center}
\includegraphics[width=0.8\linewidth]{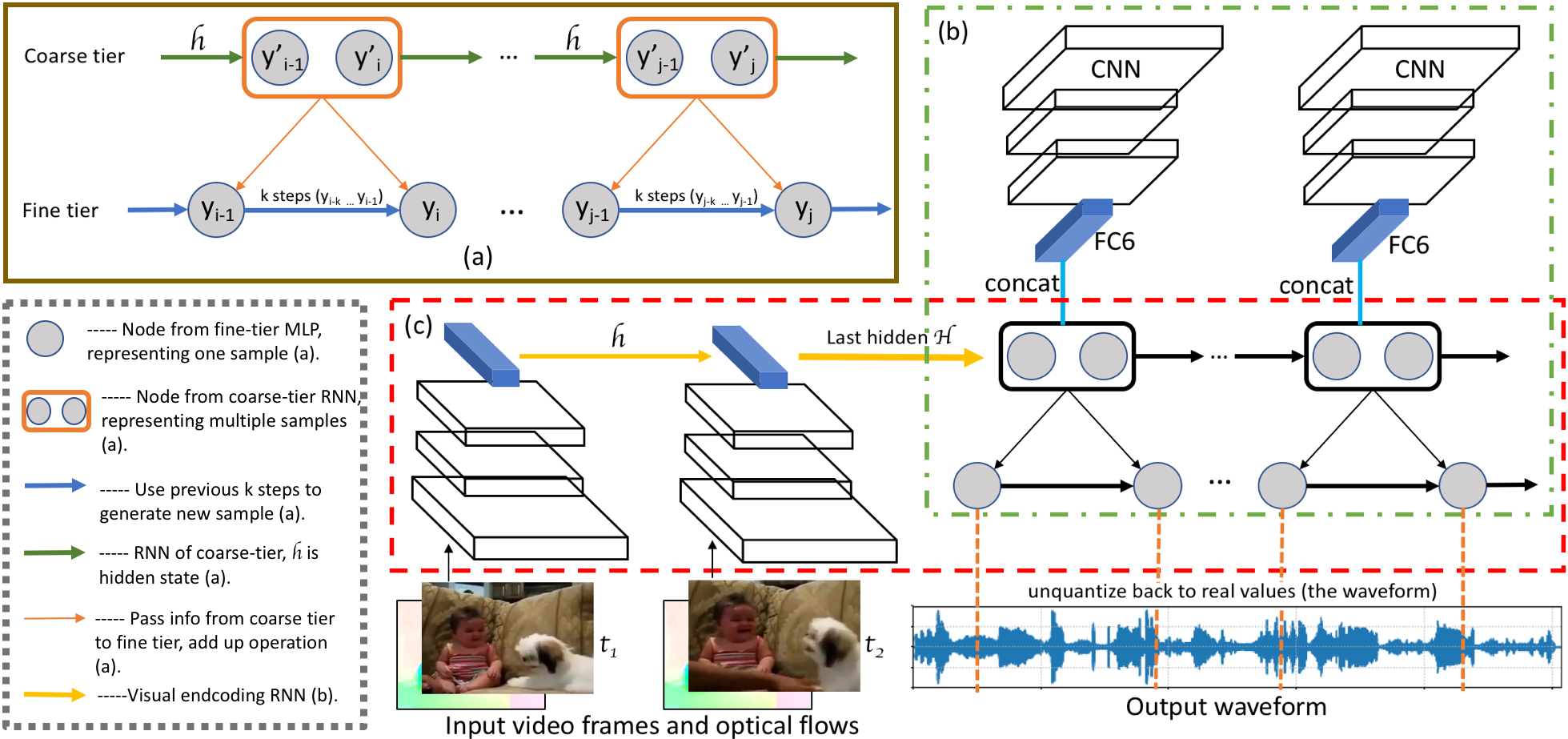}
\end{center}
   \caption{(a) (brown box) shows the simplified architecture of the sound generator, where the fine tier MLP takes as input $k$ previously generated samples and output from the coarse tier to guide generation of new samples. (b) (green dotted box) presents the frame-to-frame structure, where we concatenate the visual representation (the blue FC6 cuboid) with the nodes from the coarsest tier. And (c) (red dotted box) shows the model architecture for sequence-to-sequence and flow-based methods, we recurrently embed visual representations and use the last encoding hidden state (the bold yellow arrow) to initialize the hidden state of the coarsest tier RNN of the sound generator. The MLP tier of the sound generator does 256-way classification to output integers within $[0, 255]$, which are linearly mapped to raw waveforms $[-1, 1]$. The legends in the bottom-left gray dotted box summarize the meaning of the visualization units and the letters in the end ((a)/(b)/(c)) point to the part where the unit can be found.}
\vspace{-.3cm}
\label{fig:model_arch}
\end{figure*}

Our goal is to directly synthesize waveform samples with a generative model. As mentioned before, in order to obtain audios of reasonable quality (i.e., sounds natural), we adopt a high sampling rate at 16kHz. This requirement results in extremely long sequences, which poses challenges to a sound generator.
For this purpose, we choose the recently proposed SampleRNN~\cite{SampleRNN} as our sound generator. SampleRNN is a hierarchically structured recurrent neural network. Its coarse-to-fine structure enables the model to generate extremely long sequences and the recurrent structure of each layer captures the dependency between distant samples. SampleRNN has been applied to speech synthesis and music generation tasks previously. Here we apply it to generate natural sound for videos in the wild, which typically contain much larger variations, less structural patterns, and more noise than speech or music data.

Specifically, Fig.~\ref{fig:model_arch}(a) (upper-left corner brown box) shows the simplified overview of the SampleRNN model. Note, this simplified illustration shows 2 tiers, but more tiers are possible (we use 3). This model consists of multiple tiers, the fine tier (bottom layer) is a multilayer perceptron (MLP) which takes the output from the next coarser tier (upper layer) and the previous $k$ samples to generate a new sample. During training, the waveform samples (real numbers from -1 to 1) have been linearly quantized to integers ranging from 0 to 255, and the MLP of the finest tier can be considered a 256-way classification to predict one sample at each timestep (then mapped back to real values for the final waveform). The coarser (upper) tiers are  recurrent neural networks which can be a GRU~\cite{GRU}, LSTM~\cite{LSTM}, or any other RNN variants, and the nodes contain multiple waveform samples (2 in this illustration), meaning that this layer predicts multiple samples jointly at each time step based on previous time steps and predictions from coarser tiers. The green arrow represents the hidden state. Note that we tried using the model from WaveNet~\cite{wavenet} on the natural sound generation task, but it sometimes failed to generate meaningful sounds for categories like dog, and was outperformed by SampleRNN consistently for all object categories. Therefore, we did not pursue it further. Due to space limit, we omit the technical details of SampleRNN. For more information regarding SampleRNN, please refer to~\cite{SampleRNN}.

\subsection{Frame-to-frame method}
\label{sec:frame}
For the video encoder component, we first propose a straight-forward frame-to-frame encoding method. We represent the video frames as $x_i = V(f_i)$, where $f_i$ is the $i^{th}$ frame and $x_i$ is the corresponding representation. Here, $V(.)$ is the operation to extract the $fc6$ feature of VGG19 network~\cite{vgg19} which has been pre-trained on ImageNet~\cite{imagenet} and $x_i$ is a 4096-dimensional vector.

In this model, we encode the visual information by uniformly concatenating the frame representation with the nodes (samples) of the coarsest tier RNN of the sound generator as shown in Fig.~\ref{fig:model_arch}(b) (content in dotted green box). Due to the difference of sampling rates between the two modalities, to maintain the alignment between them, for each $x_i$, we duplicate it $s$ times, so that visual and sound sequences have the same length.
Here $s = ceiling[sr_{audio} / sr_{video}]$, where $sr_{audio}$ is the sampling rate of audio, $sr_{video}$ is that of video. Note that we only feed the visual features into the coarsest tier of SampleRNN because of the importance of this layer as it guides the generation of all finer tiers as well as for computational efficiency.

\subsection{Sequence-to-sequence method}
\label{sec:seq}
Our second model design has a sequence to sequence type of architecture \cite{seq2seq}. 
In this sequence-to-sequence model, the video encoder and sound generator are clearly separated, and connected via a bottleneck representation, which feeds encoded visual information to the sound generator. As Fig.~\ref{fig:model_arch}(c) (content in the middle red dotted box) shows, we build a recurrent neural network to encode video features. Here the same deep feature ($fc6$ layer of VGG19) is used to represent video frames as in Sec.~\ref{sec:frame}. After visual encoding (i.e., deep feature extraction and recurrent processing), we use the last hidden state from the video encoder to initialize the hidden state of the coarsest tier RNN of the sound generator, then sound generation starts. Therefore the sound generation task becomes: 
\begin{equation}
p(y_1, ..., y_n | x_1, ..., x_m) = 
\prod_{i=1}^{n} p(y_i | H, y_1, ..., y_{i-1})
\end{equation}
where $H$ represents the last hidden state of the video encoding RNN or equivalently the initial hidden state of the coarsest tier RNN of the sound generator. 

Unlike the frame based model mention above, where we explicitly enforce the alignment between video frames and waveform samples. In this sequence-to-sequence model, we expect the model to learn such alignment between the two modalities through encoding and decoding.

\subsection{Flow-based method}
\label{sec:flow}
Our third model further improves the visual representation to better capture the content and motion in input videos. As motivation for this variant, we argue that motion signals in the visual domain, even though sometimes subtle, are critical to synthesize realistic and well synchronized sound. 
For instance, the barking sound should be generated at the moment when the dog opens its month and maybe the body starts to lean forward. This requires our model to be sensitive to activities and motion of target objects. However, the previously used VGG features are pre-trained on object classification tasks, which typically result in features with rotation and translation invariance. Although the VGG features are computed along consecutive video frames, which implicitly include some motion signals, it may still fail to capture them. 
Therefore, to explicitly capture the motion signal, we add an optical flow-based deep feature to the visual encoder and call this method the flow-based method. The overall architecture of the current method is identical to the sequence-to-sequence model (as Fig.~\ref{fig:model_arch}(c) shows), which encodes video features $x_i$ recurrently through RNN and decodes with SampleRNN. The only difference is that here $x_i = cat[V(f_i) , F(o_i)]$ ($cat[.]$ indicates concatenation operation); $o_i$ is the optical flow of $i^{th}$ frame; and F(.) is the function to extract the optical flow-based deep feature. We pre-compute optical flow between video frames using \cite{flow} and feed the flows to the temporal ConvNets from \cite{two_stream}, which has been pre-trained on optical flows of UCF-101 video activity dataset \cite{ucf101}, to get the deep feature. We extract the $fc6$ layer of temporal ConvNets, a 4096-dimensional vector.

\vspace{-.2cm}
\section{Experiments}
\label{sec:experiments}

\begin{figure*}[t]
\begin{center}
\includegraphics[width=0.85\linewidth]{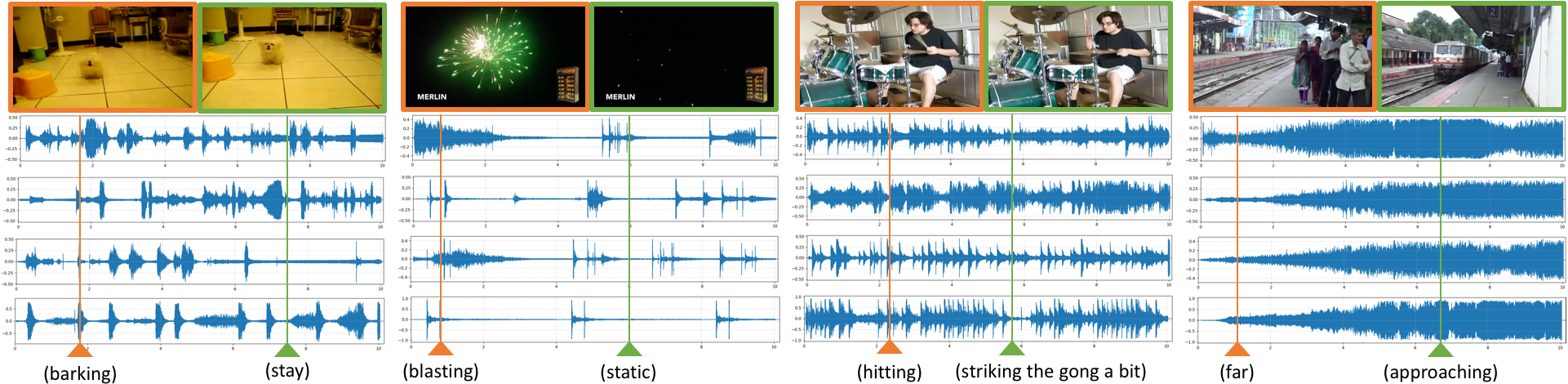}
\end{center}
   \caption{Waveforms of generated audio aligned with corresponding video key frames. From left to right showing: Dog, Fireworks, Drum, and Rail transport categories. For each case the 4 waveforms (from top to bottom) are from $Frame$, $Seq$, $Flow$ methods, and the original audio. The border color of the frames indicates which flagged position is shown and descriptions indicate what is happening in the video at that moment.}
\vspace{-.3cm}
\label{fig:results_wav}
\end{figure*}

In this section, we first introduce the model structure and training details (Sec.~\ref{sec:model_arch}). Then, we visualize the generated audio to qualitatively evaluate the results (Sec.~\ref{sec:visualization}). Quantitatively, we report the loss values for all methods and evaluate generated results on a video retrieval task (Sec.~\ref{sec:numerical_exp}). Additionally, we also run 3 human evaluation experiments to subjectively evaluate the results from the different proposed models (Sec.~\ref{sec:human_exp}).

\subsection{Model and training details}
\label{sec:model_arch}
We train the 3 proposed models on each of the 10 categories of our dataset independently. All training videos have been padded to the same length (10 secs) by duplicating and concatenating up to the target length. We sample the videos at 15.6 FPS (156 frames for 10 seconds) and sample the audios at approximately 16kHz, specifically 159744 times per 10 seconds. For the frame based method, step size $s$ is set to 1024.

\vspace{+.1cm}
\noindent {\bf Sound generator:} We apply a 3-tier SampleRNN with one-layer RNN for the coarsest and second coarsest tiers, and a MLP for the finest tier. For the finest tier, new sample generations are based on the previous $k$ generated samples ($k = 4$). We use GRU as the recurrent structure. The number of samples included by each node from coarse to fine tiers are: 8, 2 ,1 with hidden state size of 1024 for the coarsest and second coarsest tiers.

\vspace{+.1cm}
\noindent {\bf Frame-to-frame model ($Frame$):} To concatenate the visual feature (4096-D) with the nodes from the coarsest tier GRU, we first expand the node (8 samples) to 4096 by applying a fully connected operation. After combining with the visual feature, we obtain a 8192-D vector to feed into the coarsest tier of the sound generator.

\vspace{+.1cm}
\noindent {\bf Sequence-to-sequence ($Seq$) \& flow based model ($Flow$):} These two models have the same architecture. For the visual encoding recurrent neural network, we also use an one-layer-GRU structure with the hidden state size equal to 1024. The only difference is that for the flow based model, the visual feature is the concatenation of the deep image feature (4096-D) and deep flow feature (4096-D) resulting in a 8192-D vector.

\vspace{+.1cm}
We randomly select 128 videos from each category for testing, leaving the remaining videos for training. No data augmentation has been applied. During training, we apply Adam Stochastic Optimization \cite{adam} with learning rate 0.001 and minibatch of size 128 for all models. For our experiments we train models for each category independently. As an additional experiment, aiming to handle multiple audio-visual objects within the same video, we also train a multi-category model where we combine data from all categories. We show some results of the multi-category model on videos from the Internet containing multiple interacting objects in the supplementary video. 

\subsection{Qualitative visualization}
\label{sec:visualization}
We visualize the generated waveform results from the three proposed models as well as the original audio and corresponding video frames in Fig.~\ref{fig:results_wav}. Results from four categories are shown from left to right: Dog, Fireworks, Drum, and Rail transport. The former three are synchronization-sensitive categories, though that doesn't mean the waveform needs to be exactly aligned with the ground truth for good human perception. For instance in the fireworks example (second left), we show the waveforms from $Frame$, $Seq$, $Flow$ methods and the real audio from top to bottom. Compared to the real audio, the $Flow$ waveform (third) shows several extra light explosions (high peaks). When we listen to it, these extra peaks sound like far away explosions, which reasonably fits the scene. The Rail transport category is not that sensitive to the specific speed of the objects but some of the videos like the depicted example have the obvious property that the amplitude of the sound is affected by the distance of the target object (when the train approaches, the sound gets louder). All three of our models can implicitly learn this effect. We show more qualitative results in the supplementary video. 

\subsection{Numerical evaluation}
\label{sec:numerical_exp}
In this section, we provide quantitative evaluations of the models.

\vspace{+.1cm}
\noindent {\bf Loss values:} First we show the average cross-entropy loss (the finest layer of sound generator does 256-way classification for each sample prediction) for training and testing of $Frame$, $Seq$ and $Flow$ models in Table.~\ref{table:loss_value}. We can see that $Flow$ and $Seq$ methods achieve lower training and testing loss than $Frame$ method, and they are competitive. Specifically $Seq$ method has the lowest training loss after converging, while $Flow$ works best on testing loss. 

\begin{table} [t]
\begin{center}
\setlength{\tabcolsep}{1.8em} 
\scalebox{0.7}{
\begin{tabular}{|l |c |c | c |}
\cline{2-4}
\multicolumn{1}{c|}{}& Frame & Seq & Flow \\
\hline
Training  & 2.6143 & {\bf2.5991} & 2.6037 \\
Testing & 2.7061 & 2.6866 & {\bf2.6839}  \\
\hline
\end{tabular}
}
\end{center}
\caption{Average cross-entropy loss for training and testing of 3 methods. Frame represents frame-to-frame method; Seq means sequence-to-sequence method and Flow is flow based method. We mark the best results in bold.}
\vspace{-.3cm}
\label{table:loss_value}
\end{table}

\vspace{+.1cm}
\noindent {\bf Retrieval experiments:} Since direct quantitative evaluation of waveforms is quite challenging, we design a retrieval experiment that serves as a good proxy. 
Since our task is to generate audio given visual representations, well trained models should have the capability of mapping visual information to their corresponding correct (or reasonable) audio signal. To evaluate our models in this direction, we design a retrieval experiment where visual features are used as queries and audio with the maximum sampling likelihood is retrieved. Here the audios from all testing videos are combined into a database of 1280 audios, and audio-retrieval performance is measured for each testing video.

If our models have learned a reasonable mapping, the retrieved audio should be (1) from the same category as the query video (category-level retrieval), and more ideally (2) the exact audio corresponding to the query video (instance-level retrieval). Note that this can be very challenging since videos may contain very similar contents. In Table.~\ref{table:retrieval} we show the average top1 and top5 retrieval accuracy for category and instance retrieval. We observe that all methods are significantly better than chance (where chance for category retrieval is 10\% and for instance retrieval is 0.78\%). The flow based method achieves the best accuracy under both metrics.


\begin{table} [t]
\begin{center}
\setlength{\tabcolsep}{1.5em} 
\scalebox{0.65}{
\begin{tabular}{|l |c |c | c | c |}
\cline{3-5}
\multicolumn{2}{c|}{}& Frame & Seq & Flow \\
\hline
Category & Top1 & 40.55\% & 44.14\% & {\bf45.47\%} \\
\cline{2-5}
  & Top5 & 53.59\% & 58.28\% & {\bf60.31\%} \\
\cline{1-5}
Instance & Top1 & 4.77\% & 5.70\% & {\bf5.94\%} \\
\cline{2-5}
 & Top5 & 7.81\% & 9.14\% & {\bf10.08\%} \\
\hline
\end{tabular}
}
\end{center}
\caption{Top 1 and top 5 audio retrieval accuracy. 'Category' measures category-level retrieval, while 'Instance' indicates instance-level retrieval.}
\vspace{-.4cm}
\label{table:retrieval}
\end{table}

\subsection{Human evaluation experiments}
\label{sec:human_exp}
As in image or video generation tasks, the quality of generated results can be very subjective. For instance, sometimes even though the generated sound or waveforms might not be very similar to the ground truth (the real sound), the generation may still sound like a reasonable match to the video. This is especially true for ambient sound categories (e.g. water flow, printer) where the overall pattern may be more important than the specific frequencies, etc. Thus, comparing generated sounds with ground truth by applying distance metrics might not be the ideal way to evaluate quality. In this section, we directly compare the sound generation results from each proposed method in three human evaluation experiments on AMT. 

\vspace{+.1cm}
\noindent {\bf Methods comparison task:} This task aims to directly compare the sounds generated by the three proposed methods in a forced-choice evaluation. For each test video, we show turkers the video with audio generated by: the $Frame$, $Seq$ and $Flow$ methods. Turkers are posed with four questions and asked to select the best video-audio pair for each question. Questions are related to: 1) the correctness of the generated sound (which one sounds most likely to come from the visual contents); 2) which contains the least irritating noise; 3) which is best temporally synchronized with the video; 4) which they prefer overall. 
Each question for each test video has been labeled by 3 different turkers and we aggregate their votes to get the final results. 

\begin{figure}[t]
\begin{center}
\includegraphics[width=0.92\linewidth]{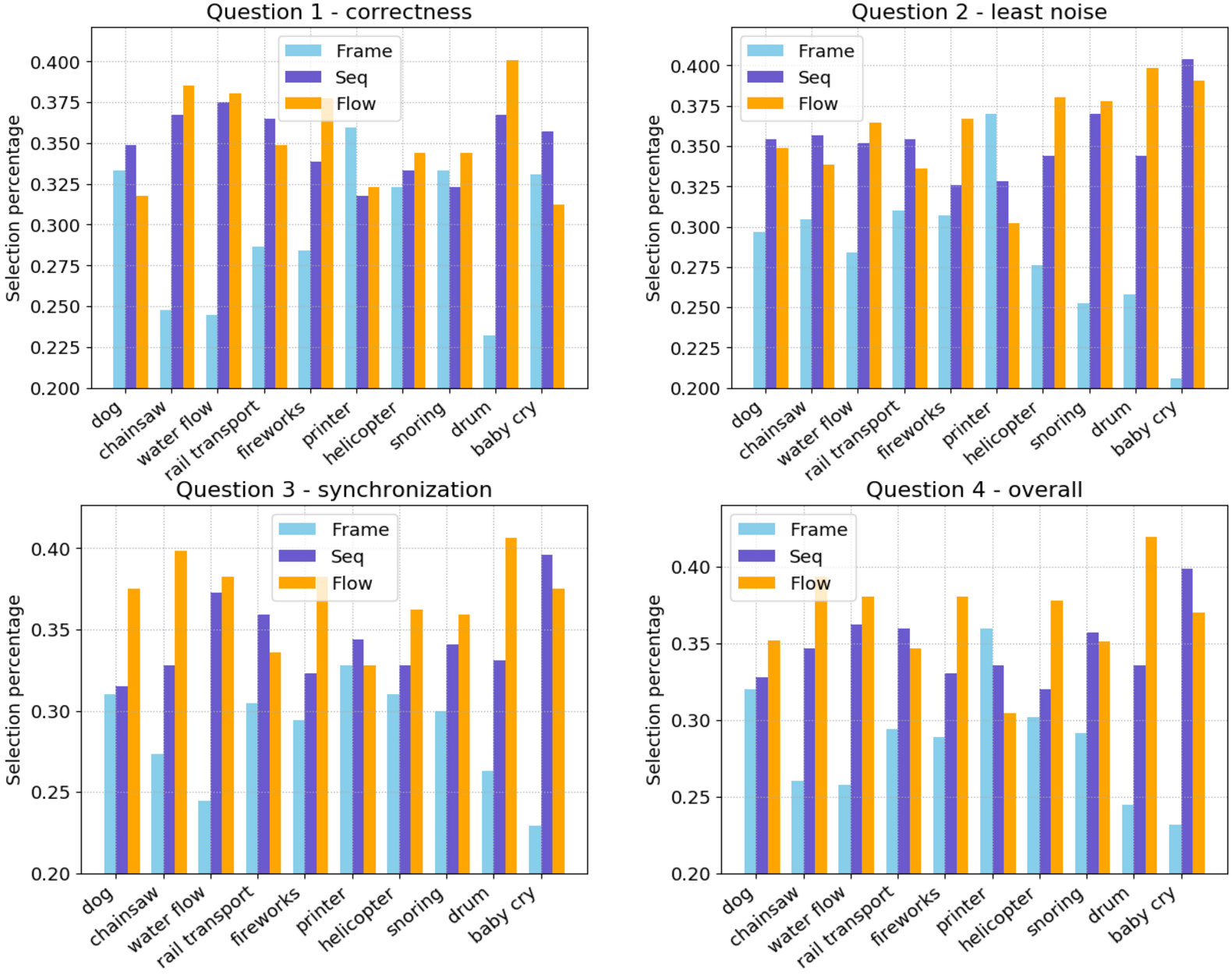}
\end{center}
\caption{Human evaluation of forced-choice experiments for four questions broken down by category.}
\vspace{-.4cm}
\label{fig:human_1}
\end{figure}

Table.~\ref{table:humanexp_forced_sel} shows the average preference rate for all categories (the higher the better) on each question. We can see that both $Seq$ and $Flow$ outperform $Frame$ based method with $Flow$ performing best overall. $Flow$ outperforms $Seq$ the most on question 3 which demonstrates that adding the deep flow feature helps with improving the temporal synchronization of visual and sound during generation. Fig.~\ref{fig:human_1} shows the results for each category. We observe that the advantage of the $Flow$ method is mainly gained on categories that are sensitive to synchronization, such as Fireworks and Drum (see question 3 and question 2 histograms).

\begin{table} [t]
\begin{center}
\setlength{\tabcolsep}{1.5em} 
\scalebox{0.65}{
\begin{tabular}{|l |c |c | c |}
\cline{2-4}
\multicolumn{1}{c|}{}& Frame & Seq & Flow \\
\hline
Correctness  & 29.74\% & 34.92\% & {\bf35.34\%} \\
Least noise & 28.65\% & 35.31\% & {\bf36.04\%}  \\
Synchronization & 28.57\% & 34.37\% & {\bf37.06\%}  \\
Overall & 28.52\% & 34.74\% & {\bf36.74\%}  \\
\hline
\end{tabular}
}
\end{center}
\caption{Human evaluation results in a forced-choice selection task. Here we show the average selection rate percentage over all categories for each of 4 questions.}
\label{table:humanexp_forced_sel}
\end{table}

\vspace{+.1cm}
\noindent {\bf Visual relevance task:}
Synchronization between video and audio can be one of the fundamental factors to measure the realness of sound generated from videos, but synchronization tolerance can vary between categories. 
For example, we easily detect discordance when a barking sound fails to align to the correct barking motion of a dog. While for other categories like water flowing, we might be more tolerant, and may not notice if we swapped the audio from one river to a another.
Inspired by this observation, we design a task in which each test video is combined with two audios, and ask the turkers to pick the video-audio pair that best corresponds. One of the audios is generated from the video, while the other is randomly chosen from another video of the same category.
This task measures whether the audio we generate is discriminative for the input video.

Table.~\ref{table:humanexp_best_fit} shows the percentage of matched audios being correctly selected. Results are reported for the sounds generated by our three methods (the first three columns) as well as the real sounds (the last column). Each test sample is rated by three turkers.
The results are consistent with our intuition -- real sounds are generally discriminative for the corresponding videos, though some categories like helicopter or water are less discriminative. Our three methods achieve reasonable accuracy in some categories like Dog and Fireworks (outperforming 50\% chance by a large margin). For more ambient sound categories, the discrimination task is challenging for both generated and real audio.


\begin{table} [t]
\begin{center}
\setlength{\tabcolsep}{1.4em} 
\scalebox{0.62}{
\begin{tabular}{|l |c |c | c || c |}
\cline{2-5}
\multicolumn{1}{c|}{}& Frame & Seq & Flow & Real\\
\hline
Dog  & 57.29\% & 58.85\% & {\bf63.02\%} & 75.00\% \\
Chainsaw & 56.25\% & 57.03\% & {\bf58.07\%} & 70.31\% \\
Water flowing & 49.22\% & 52.34\% & {\bf52.86\%} & 59.90\% \\
Rail transport & 53.39\% & {\bf56.51\%} & 55.47\% & 66.14\% \\
Fireworks & 61.98\% & 67.97\% & {\bf68.75\%} & 79.17\% \\
Printer & 46.09\% & {\bf50.52\%} & 47.14\% & 60.16\% \\
Helicopter & 51.82\% & {\bf54.95\%} & 54.17\% & 58.33\% \\
Snoring & 51.82\% & 53.65\% & {\bf54.95\%} & 63.02\% \\
Drum & 55.21\% & 59.38\% & {\bf62.24\%} & 73.44\% \\
Baby crying & 52.60\% & {\bf57.55\%} & 56.77\% & 70.57\% \\
\hline
Average & 53.56\% & 56.88\% & {\bf57.34\%} & 67.60\% \\
\hline
\end{tabular}
}
\end{center}
\caption{Human evaluation results: visual relevance. Rows show the selection accuracy for each category and their average. `Real' stands for using original audios.}
\vspace{-.2cm}
\label{table:humanexp_best_fit}
\end{table}

\vspace{+.1cm}
\noindent {\bf Real or fake determination:} In this task, we would like to see whether the generated audios can fool people into thinking that they are real. We provide instructions to the turkers that the audio of the current video might be either real (originally belonging to this video) or fake (synthesis by computers). The criteria of being fake can be bad synchronization or poor quality such as containing unpleasing noise. In addition to the generated results from our proposed methods, we also include videos with the original audio as a control. As an additional baseline, we also combine the video with a random real audio from the same category. This baseline is rather challenging as it uses real audios. Each evaluation is performed by 3 turkers and we aggregate the votes. 

The percentages for the audios being rated as real are shown in Table.~\ref{table:humanexp_real_fake} for all methods including the baseline ($Base$) and the real audio. $Seq$ and $Flow$ methods outperform the $Frame$ method except for the printer category. Unsurprisingly, $Base$ achieves decent results on categories that are insensitive to synchronization like Printer and Snoring, but much worse than our methods on categories sensitive to synchronization such as Dog and Drum. One of the reasons that turkers consider some of the real cases as fake is that a few original audios might include light background music or other noise which appears not fitting with the visual content.

\begin{table} [t]
\begin{center}
\setlength{\tabcolsep}{0.92em} 
\scalebox{0.62}{
\begin{tabular}{|l |c |c | c || c | c|}
\cline{2-6}
\multicolumn{1}{c|}{}& Frame & Seq & Flow & Base& Real\\
\hline
Dog  & 61.46\% & {\bf64.32\%} & 62.24\% & 54.69\% & 89.06\% \\
Chainsaw & 71.09\% & 73.96\% & {\bf76.56\%} & 68.23\% & 93.75\% \\
Water flowing & 70.83\% & 77.60\% & {\bf81.25\%} & 77.86\% & 87.50\% \\
Rail transport & 79.69\% & {\bf83.33\%} & 80.47\% & 74.74\% & 90.36\% \\
Fireworks & 76.04\% & 76.82\% & {\bf78.39\%} & 75.78\% & 94.01\% \\
Printer & 73.96\% & 73.44\% & 71.35\% & {\bf75.00\%} & 89.32\% \\
Helicopter & 71.61\% & 74.48\% & 78.13\% & {\bf78.39\%} & 91.67\% \\
Snoring & 67.71\% & 73.44\% & 73.18\% & {\bf77.08\%} & 90.63\% \\
Drum & 62.24\% & 64.58\% & {\bf70.83\%} & 59.64\% & 93.23\% \\
Baby crying & 57.29\% & 64.32\% & 61.20\% & {\bf69.27\%} & 94.79\% \\
\hline
Average & 68.69\% & 72.63\% & {\bf73.36\%} & 71.07\% & 91.43\% \\
\hline
\end{tabular}
}
\end{center}
\caption{Human evaluation results: real or fake task where people judge whether a video-audio pair is real or generated. Percentages indicate the frequency of a pair being judged as real.}
\vspace{-.5cm}
\label{table:humanexp_real_fake}
\end{table}

\subsection{Additional experiments}

\noindent {\bf Multi-category results:} We test our multi-category model on the VEGAS dataset by conducting the real/fake experiment in Sec~\ref{sec:human_exp} and find on average 46.29\% of the generated sound can fool human (versus 73.63\% of the best single-category model). Note a random baseline is virtually 0\% as humans are very sensitive to sound. Another solution of multi-category results can be achieved by utilizing the state-of-the-art visual classification algorithms to get the category label before applying the per-category models.

\vspace{+.1cm}
\noindent {\bf Comparison with \cite{visual_indicated_sound}:} \cite{visual_indicated_sound} presents a CNN stacked with RNN structure to predict sound features (cochleagrams) at each time step, and audio samples are reconstructed by example-based retrieval. We implement an upper bound version by assuming the cochleagrams of ground truth sound are given for test videos. And we retrieve the sound from training data with the stride of 2s. This provides a baseline stronger than the method in~\cite{visual_indicated_sound}. We do not observe noticeable artifacts on the boundary of retrieved sound segments, but the synthesized audio does not synchronized very well with the visual content. We also conduct the same real/fake evaluation on the Dog and Drum categories, and the generated sound with this upper bound can fool 40.16\% and 43.75\% of human subjects respectively, which are largely outperformed by our results (64.32\% and 70.83\%).

On the other hand, we also test our model on the Greatest Hits dataset from \cite{visual_indicated_sound}. Note that our model has been trained to generate much longer audios (10s) than those in \cite{visual_indicated_sound} (0.5s). We evaluate the model via a similar psychology study as described in Sec~6.2 of \cite{visual_indicated_sound}. 41.50\% of our generated sounds are favored by humans over real sound, which is competitive with the method in \cite{visual_indicated_sound} that achieves 40.01\%. The experiments show the generalization capability of our model.

\vspace{-.2cm}
\section{Conclusion}
In this work, we introduced the task of generating realistic sound from videos in the wild. We created a dataset for this purpose, sampled from the AudioSet collection, based on which we trained three different visual-to-sound deep network variants. We also provided qualitative, quantitative and subjective experiments to evaluate the models and the generated audio results. Evaluations show that over 70\% of the generated sound from our models can fool humans into thinking that they are real. Future directions include explicitly recognizing and reasoning about objects in the video during sound generation, and reasoning beyond the pixels and temporal duration of the input frames for more contextual generation.

\vspace{+.1cm}
\noindent {\bf Acknowledgments:} This work was supported by NSF Grants \#1633295, 1562098 and 1405822.

{\small
\bibliographystyle{ieee}
\bibliography{egbib}

\begin{thebibliography}{10}\itemsep=-1pt

\bibitem{Youtube8M}
S.~Abu{-}El{-}Haija, N.~Kothari, J.~Lee, P.~Natsev, G.~Toderici,
  B.~Varadarajan, and S.~Vijayanarasimhan.
\newblock Youtube-8m: {A} large-scale video classification benchmark.
\newblock {\em CoRR}, 2016.

\bibitem{look}
R.~Arandjelovi\'c and A.~Zisserman.
\newblock Look, listen and learn.
\newblock In {\em ICCV}, 2017.

\bibitem{soundnet}
Y.~Aytar, C.~Vondrick, and A.~Torralba.
\newblock Soundnet: Learning sound representations from unlabeled video.
\newblock In {\em NIPS}, 2016.

\bibitem{v2s2v}
L.~Chen, S.~Srivastava, Z.~Duan, and C.~Xu.
\newblock Deep cross-modal audio-visual generation.
\newblock {\em CoRR}, 2017.

\bibitem{GRU}
K.~Cho, B.~van Merrienboer, Çaglar G\"{u}lçehre, D.~Bahdanau, F.~Bougares,
  H.~Schwenk, and Y.~Bengio.
\newblock Learning phrase representations using rnn encoder-decoder for
  statistical machine translation.
\newblock In {\em EMNLP}. 2014.

\bibitem{imagenet}
J.~Deng, W.~Dong, R.~Socher, L.-J. Li, K.~Li, and L.~Fei-Fei.
\newblock Imagenet: A large-scale hierarchical image database.
\newblock In {\em CVPR}, 2009.

\bibitem{audioset}
J.~F. Gemmeke, D.~P.~W. Ellis, D.~Freedman, A.~Jansen, W.~Lawrence, R.~C.
  Moore, M.~Plakal, and M.~Ritter.
\newblock Audio set: An ontology and human-labeled dataset for audio events.
\newblock In {\em ICASSP}, 2017.

\bibitem{GAN}
I.~Goodfellow, J.~Pouget-Abadie, M.~Mirza, B.~Xu, D.~Warde-Farley, S.~Ozair,
  A.~Courville, and Y.~Bengio.
\newblock Generative adversarial nets.
\newblock In {\em NIPS}. 2014.

\bibitem{spoken}
D.~Harwath, A.~Torralba, and J.~R. Glass.
\newblock Unsupervised learning of spoken language with visual context.
\newblock In {\em NIPS}, 2016.

\bibitem{LSTM}
S.~Hochreiter and J.~Schmidhuber.
\newblock Long short-term memory.
\newblock {\em Neural Comput.}, 1997.

\bibitem{unit_tts}
A.~Hunt and A.~Black.
\newblock Unit selection in a concatenative speech synthesis system using a
  large speech database.
\newblock In {\em ICASSP}, 1996.

\bibitem{sports-1M}
A.~Karpathy, G.~Toderici, S.~Shetty, T.~Leung, R.~Sukthankar, and L.~Fei-Fei.
\newblock Large-scale video classification with convolutional neural networks.
\newblock In {\em CVPR}, 2014.

\bibitem{adam}
D.~P. Kingma and J.~Ba.
\newblock Adam: {A} method for stochastic optimization.
\newblock {\em ICLR}, 2014.

\bibitem{SampleRNN}
S.~Mehri, K.~Kumar, I.~Gulrajani, R.~Kumar, S.~Jain, J.~Sotelo, A.~C.
  Courville, and Y.~Bengio.
\newblock Samplernn: An unconditional end-to-end neural audio generation model.
\newblock {\em ICLR}, 2016.

\bibitem{visual_indicated_sound}
A.~Owens, P.~Isola, J.~McDermott, A.~Torralba, E.~Adelson, and W.~Freeman.
\newblock Visually indicated sounds.
\newblock In {\em CVPR}, 2016.

\bibitem{ambient}
A.~Owens, J.~Wu, J.~H. McDermott, W.~T. Freeman, and A.~Torralba.
\newblock Ambient sound provides supervision for visual learning.
\newblock In {\em ECCV}, 2016.

\bibitem{two_stream}
K.~Simonyan and A.~Zisserman.
\newblock Two-stream convolutional networks for action recognition in videos.
\newblock In {\em NIPS}. 2014.

\bibitem{vgg19}
K.~Simonyan and A.~Zisserman.
\newblock Very deep convolutional networks for large-scale image recognition.
\newblock {\em ICLR}, 2015.

\bibitem{ucf101}
K.~Soomro, A.~R. Zamir, M.~Shah, K.~Soomro, A.~R. Zamir, and M.~Shah.
\newblock Ucf101: A dataset of 101 human actions classes from videos in the
  wild.
\newblock {\em CoRR}, 2012.

\bibitem{Char2Wav}
J.~Sotelo, S.~Mehri, K.~Kumar, J.~F. Santos, K.~Kastner, A.~Courville, and
  Y.~Bengio.
\newblock Char2wav: End-to-end speech synthesis.
\newblock {\em ICLR}, 2017.

\bibitem{flow}
D.~Sun, S.~Roth, and M.~J. Black.
\newblock Secrets of optical flow estimation and their principles.
\newblock In {\em CVPR}, 2010.

\bibitem{seq2seq}
I.~Sutskever, O.~Vinyals, and Q.~V. Le.
\newblock Sequence to sequence learning with neural networks.
\newblock NIPS, 2014.

\bibitem{wavenet}
A.~van~den Oord, S.~Dieleman, H.~Zen, K.~Simonyan, O.~Vinyals, A.~Graves,
  N.~Kalchbrenner, A.~W. Senior, and K.~Kavukcuoglu.
\newblock Wavenet: {A} generative model for raw audio.
\newblock {\em CoRR}, 2016.

\bibitem{Robust_tts}
J.~Yamagishi, T.~Nose, H.~Zen, Z.~H. Ling, T.~Toda, K.~Tokuda, S.~King, and
  S.~Renals.
\newblock Robust speaker-adaptive hmm-based text-to-speech synthesis.
\newblock {\em IEEE Transactions on Audio, Speech, and Language Processing},
  2009.

\bibitem{hmm_tts}
T.~Yoshimura, K.~Tokuda, T.~Masuko, T.~Kobayashi, and T.~Kitamura.
\newblock Simultaneous modeling of spectrum, pitch and duration in hmm-based
  speech synthesis.
\newblock In {\em Eurospeech}, 1999.

\bibitem{dnn_tts}
H.~Zen, A.~Senior, and M.~Schuster.
\newblock Statistical parametric speech synthesis using deep neural networks.
\newblock In {\em ICASSP}, 2013.

\bibitem{stat_tts}
H.~Zen, K.~Tokuda, and A.~W. Black.
\newblock Statistical parametric speech synthesis.
\newblock {\em Speech Communication}, 2009.

\bibitem{sound20k}
Z.~Zhang, J.~Wu, Q.~Li, Z.~Huang, J.~Traer, J.~H. McDermott, J.~B. Tenenbaum,
  and W.~T. Freeman.
\newblock Generative modeling of audible shapes for object perception.
\newblock In {\em ICCV}, 2017.

\end{thebibliography}
}

\end{document}